\documentclass[5p,times,twocolumn]{elsarticle}

\usepackage{amsmath,amssymb,amsfonts,color}
\usepackage{graphicx}
\usepackage{textcomp}
\usepackage{xcolor}
\usepackage{enumerate}
\usepackage{setspace}
\usepackage{bm}
\usepackage{comment}
\usepackage{caption}
\usepackage{tipa}
\usepackage{footnote}
\usepackage{subfigure}
\usepackage{algorithm}
\usepackage{algorithmic}
\usepackage{url}

\def\x{{\mathbf x}}

\begin{document}

\begin{frontmatter}

\title{Compressed Gastric Image Generation Based on Soft-Label Dataset Distillation for \\ Medical Data Sharing}

\author{Guang Li${}^\text{a}$}
\ead{guang@lmd.ist.hokudai.ac.jp}
\author{Ren Togo${}^\text{b}$}
\ead{togo@lmd.ist.hokudai.ac.jp}
\author{Takahiro Ogawa${}^\text{b}$}
\ead{ogawa@lmd.ist.hokudai.ac.jp}
\author{Miki Haseyama${}^\text{b}$}
\ead{mhaseyama@lmd.ist.hokudai.ac.jp}
\address{${}^\text{a}$Graduate School of Information Science and Technology, Hokkaido University, \\
           N-14, W-9, Kita-Ku, Sapporo, 060-0814, Japan}
\address{${}^\text{b}$Faculty of Information Science and Technology, Hokkaido University, \\
           N-14, W-9, Kita-Ku, Sapporo, 060-0814, Japan}

\begin{abstract}

{\it Background and objective:~} Sharing of medical data is required to enable the cross-agency flow of healthcare information and construct high-accuracy computer-aided diagnosis systems.
However, the large sizes of medical datasets, the massive amount of memory of saved deep convolutional neural network (DCNN) models, and patients$'$ privacy protection are problems that can lead to inefficient medical data sharing.
Therefore, this study proposes a novel soft-label dataset distillation method for medical data sharing.

{\it Methods:~} 
The proposed method distills valid information of medical image data and generates several compressed images with different data distributions for anonymous medical data sharing.
Furthermore, our method can extract essential weights of DCNN models to reduce the memory required to save trained models for efficient medical data sharing. 

{\it Results:~} The proposed method can compress tens of thousands of images into several soft-label images and reduce the size of a trained model to a few hundredths of its original size.
The compressed images obtained after distillation have been visually anonymized; therefore, they do not contain the private information of the patients.
Furthermore, we can realize high-detection performance with a small number of compressed images.

{\it Conclusions:~} The experimental results show that the proposed method can improve the efficiency and security of medical data sharing.

\end{abstract}


\end{frontmatter}

\section{Introduction}
\label{sec1}
Healthcare refers to practical measures for maintaining and promoting of physical and mental health, including standards for preventing mental illnesses and physical diseases caused by work, life, environment, and psychological factors~\cite{yang2015hybrid}.
Generally, quality healthcare services are provided by identifying health problems, introducing innovative solutions, and allocating health resources~\cite{ali2017multi}.
Thus, these medical services focus on properly collecting, managing, and using healthcare information~\cite{schiza2018proposal, xie2018user, soni2020state}. 
Although a large medical facility may provide adequate healthcare information, a single small medical facility does not have the capacity.
Considering the insufficient amount of healthcare information in a single small medical facility, it is necessary to obtain adequate healthcare information through transmission between multiple medical facilities~\cite{esmaeilzadeh2020impact, ye2021management}.
The sharing of medical data plays an essential role in enabling the cross-agency flow of healthcare information; thus, improving the quality of medical services~\cite{dankar2017risk}.
\par
With the development of deep convolutional neural networks (DCNNs)~\cite{krizhevsky2012imagenet, lecun2015deep}, DCNN-based computer-aided diagnosis (CAD) systems~\cite{shin2016deep, litjens2017survey} have made significant progress in the healthcare field.
Since obtaining high-quality medical image data and annotating it by experienced physicians is time-consuming and expensive, the amount of data from a small medical facility is insufficient to train high-accuracy DCNNs~\cite{luo2016big}.
Therefore, sharing medical data from different medical facilities, including image data and trained DCNN models, is essential.
However, the large size of medical datasets, the massive amount of memory of saved DCNN models, and patients$'$ privacy protection are significant obstacles in medical data sharing.
With the increase in high-resolution medical image data, the sizes of medical datasets have increased exponentially~\cite{tsymbal2014towards}. 
Since there are many parameters in DCNN models, a massive amount of memory is required to save the trained DCNN models~\cite{sohoni2019low}.
Furthermore, medical images often contain private information of patients, and the use of these medical data remains controversial~\cite{narendra2016medical}.
Therefore, it is necessary to improve the efficiency and security of medical data sharing.
\par
Many methods for reducing the sizes of datasets and saved DCNN models have been proposed.
Conventional dataset reduction methods aim to select a subset and achieve performance comparable to the original dataset~\cite{olvera2010review, bachem2017practical}.
Additionally, many researchers have proposed approaches to prune networks to solve the problem of a large amount of memory required for saved DCNN models.
Network pruning methods aim to prune redundant weights and only keep essential weights.
A typical procedure for network pruning consists of training a large model, pruning the trained model, and fine-tuning the pruned model~\cite{liu2019rethinking}.
\par
Although dataset reduction and network pruning methods can be used to address inefficiencies in medical data sharing, they still have limitations.   
First, dataset reduction methods can be used to reduce the sizes of medical datasets and improve the efficiency of medical data sharing. Meanwhile, these methods can only reduce the size of a dataset to a certain extent since they use actual data from the original dataset, especially in the case of high-resolution and complex medical datasets.
Second, we can use model pruning methods to reduce the sizes of trained DCNNs in medical datasets, which can also improve the efficiency of medical data sharing.
However, all pruning methods require deleting the network architecture (i.e., channels or layers) or the use of dedicated hardware/libraries, which are not flexible and not sufficiently straightforward~\cite{liu2019rethinking}.
\par
The issue of privacy protection is also a significant barrier to medical data sharing~\cite{gkoulalas2014publishing}.
Privacy information in medical image data causes many medical facilities to be reluctant to share medical data, which has hindered research on CAD systems~\cite{mcgraw2009privacy, malin2010technical}. 
The ground truth of medical image datasets is often associated with personal information, e.g., the patient$'$s name, gender, age, and ID number.
Some methods for removing or encrypting the patient$'$s personal information have been proposed to protect the privacy of patients~\cite{bouzelat1996extraction, loukides2010anonymization, khedr2017securemed}.
However, medical images still contain patient$'$s information, such as physical characteristics and condition of the patient, which is challenging to share with different medical facilities or taken out of hospitals.
To the best of our knowledge, there is currently no way to achieve adequate anonymization of medical images.
\par
Gastric X-ray image data has the above-described problems in medical data sharing.
Furthermore, gastric cancer is the third leading cause of cancer-related death worldwide, with half of the total global gastric cancer cases occurring in Eastern Asia~\cite{ferlay2015cancer}.
It has been revealed that gastritis is the leading risk factor for the onset of gastric cancer and can be diagnosed by gastric X-ray images in an X-ray examination~\cite{uemura2001helicobacter, ohata2004progression}. 
Since gastric X-ray image data sharing is crucial for effectively constructing CAD systems, and we have previously proposed methods for automatic detection of gastritis~\cite{kanai2019gastritis, togo2019detection}, we focused on gastric X-ray images in this study. 
Although our study focused on gastric X-ray image data, we believe the findings and methodology could also be applied to other medical images, such as endoscopic and Computed Tomography (CT) images.
\par
In this paper, we propose a compressed gastric image generation method based on the soft-label dataset distillation for efficient anonymous medical data sharing.
Using the proposed method, we can distill valid information of the entire gastric X-ray image dataset into several soft-label compressed images and effectively compress the sizes of saved models.
Since compressed images are not generated from the actual distribution of the original dataset, the generation of compressed images provides a possibility for the visual anonymization of medical images.
Figure~\ref{fig0} shows an overview of the proposed method.
Although the dataset distillation~\cite{wang2018dataset} method has been proposed for distilling information of simple datasets (i.e., MNIST and CIFAR10), its use for large-size and high-resolution medical datasets has not yet been explored.
Unlike conventional dataset reduction methods, our method aims to distill information from a large dataset into several soft-label compressed anonymous images with different data distributions.
Thus, a smaller amount of data is required to learn the same amount of information as traditional dataset reduction methods.
Furthermore, our method can reduce the memory required to save trained models by extracting essential weights straightforwardly.
\par
The main contributions of this study are summarized as follows:
\begin{figure}[t]
        \centering
        \includegraphics[width=8cm]{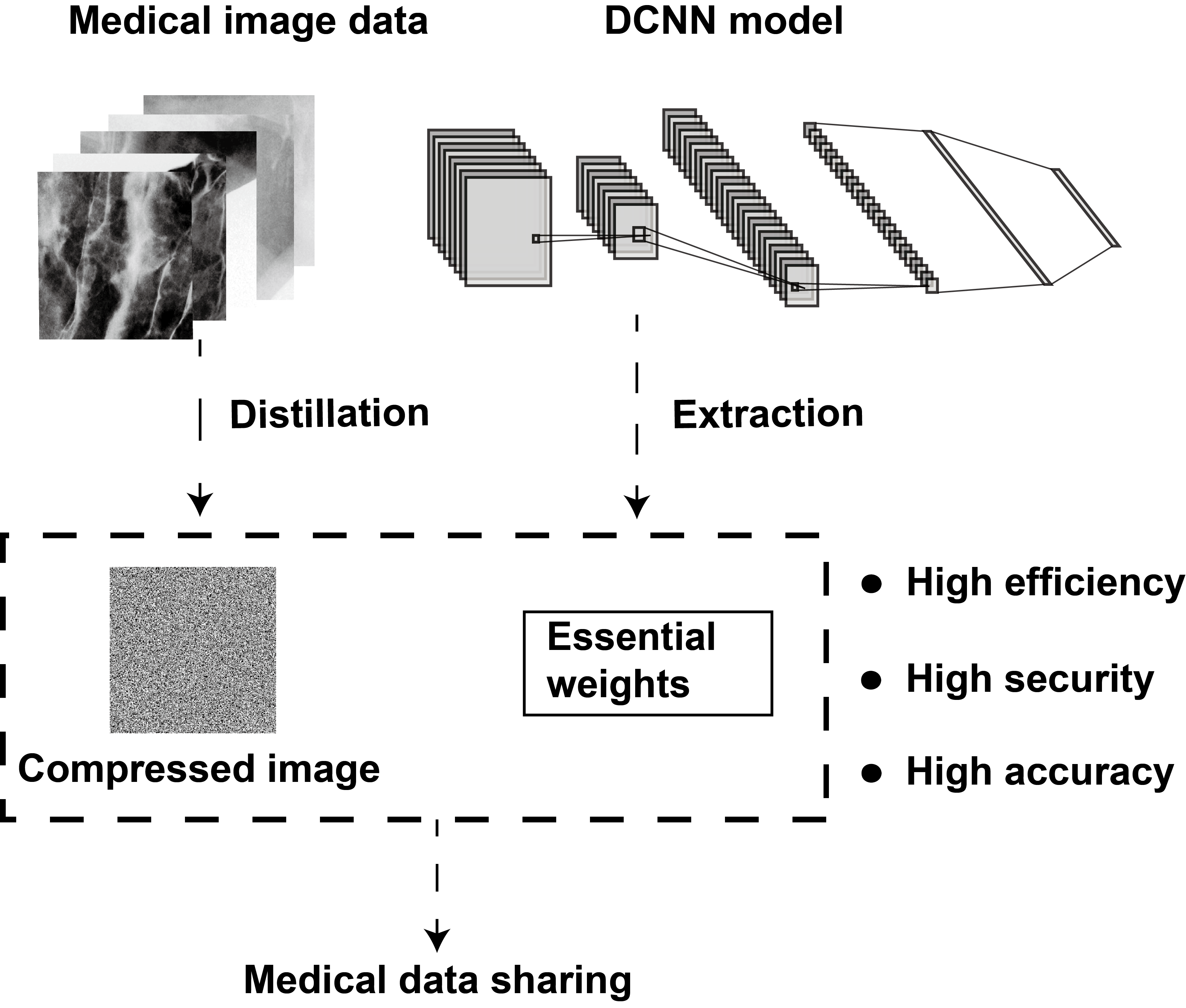}
        \caption{Overview of the proposed method.}
        \label{fig0}
\end{figure}
\begin{itemize}
    \item The proposed method can compress tens of thousands of images into several soft-label images and reduce the size of a trained model to a few hundredths of its original size.
    \item The compressed images obtained after distillation have been visually anonymized; therefore, they do not contain private information of the patients.
    \item High-detection performance can be realized with a small number of compressed images.
\end{itemize}
\par
Note that some preliminary results of this work have been presented in our previous study~\cite{li2020soft}, where we showed the feasibility of dataset distillation for gastric X-ray image data with three compressed images (one image per category).
In this paper, we extend our previous work as follows. First, we show the feasibility of distilling information of the entire gastric X-ray image dataset into only one compressed anonymous soft-label patch image for the maximum compression rate. 
Second, we demonstrate how to reduce the memory required to save trained models with batch normalization layers.
Third, we evaluate the performance of different DCNN models to verify the robustness of our method.
Finally, we discuss the relationships between the minimum number of compressed images of different DCNN models and their parameters.
\par
The remainder of this paper is organized as follows. 
In Section~\ref{sec2}, we introduce related works. 
In Section~\ref{sec3}, we describe the algorithm of the proposed method.
Then, we present the experiments, discussion, and conclusion in Sections~\ref{sec4},~\ref{sec5}, and~\ref{sec6}, respectively.
\section{Related works}
\label{sec2}
This section presents some dataset reduction methods in~\ref{sec2-A}.
Then, we review some network pruning methods in~\ref{sec2-B}.
Finally, we show some privacy protection methods in~\ref{sec2-C}.
\subsection{Dataset reduction}
\label{sec2-A}
Many methods have been proposed for making a model trained on a reduced small dataset perform as well as a model trained on the full dataset. 
Some methods create a core-set or prune the original full dataset by only using valuable data for model training, a process called core-set construction~\cite{tsang2005core, har2007smaller, bachem2017practical, campbell2018bayesian} or instance selection~\cite{angelova2005pruning, felzenszwalb2009object, olvera2010review, lapedriza2013all}.
Additionally, some methods based on active learning aim to reduce the data volume that needs to be labeled by only annotating data that are difficult to recognize~\cite{cohn1996active, tong2001support, sener2018active}.
\par
Unlike these dataset reduction methods that use actual data from the original dataset, the dataset distillation method aims to distill information from a large dataset and generate several compressed images with different distributions.   
Therefore, the dataset distillation method requires a smaller amount of data to learn the same amount of information than the amount of data required by traditional dataset reduction methods.     
However, the efficacy of the proposed dataset distillation method has only been tentatively validated for simple datasets such as MNIST and CIFAR10).
Considering the inefficiencies associated with the sharing of large-scale medical datasets, we propose the soft-label dataset distillation method for medical images and provide the first demonstration of its effectiveness for a complex gastric X-ray image dataset.
\subsection{Network pruning}
\label{sec2-B}
Over-parameterization is a universally recognized characteristic of DCNNs, and it leads to high computational costs of inference and large storage footprints for trained models~\cite{denton2014exploiting, ba2014deep}.
Recently, many network pruning methods have been proposed for pruning redundant weights and only keeping essential weights, which can reduce the computational costs and memory required to save the trained models~\cite{liu2019rethinking}.  
For example, Han \textit{et al.} proposed the deep compression pipeline: pruning, trained quantization, and Huffman coding to obtain highly compressed models without affecting their accuracy~\cite{han2015learning, han2016deep, han2016eie}.
Hinton \textit{et al.} proposed a model compression method by distilling information from large networks to a single compact network~\cite{hinton2014distilling}. 
Other network pruning methods that prune at the level of layers or channels have also been proposed~\cite{liu2017learning, luo2017thinet, molchanov2017variational}.
\par
Although these methods have good results in network pruning, they require the use of dedicated hardware/libraries or deleting the original network architecture (i.e., channels or layers)~\cite{liu2019rethinking}.
However, we found that the dataset distillation method can reduce the sizes of datasets and compress the sizes of trained models with no additional operation being required.
This paper also presents new findings on the compression of the sizes of trained models.
\subsection{Privacy protection}
\label{sec2-C}
Privacy protection issues have been hindering the development of medical data sharing~\cite{mcgraw2009privacy, malin2010technical}.
Several researchers have addressed some of the issues with anonymization.
For example, some methods for removing or encrypting the patient$'$s personal information have been proposed to protect the privacy of patients~\cite{bouzelat1996extraction, loukides2010anonymization, khedr2017securemed}.
Additionally, some methods based on blockchain~\cite{azaria2016medrec, fan2018medblock, chen2021blockchain, tan2021towards, kumar2021integration} and cloud computing~\cite{van2013lifelong, alshagathrh2018building, doel2017gift, bao2021efficient, sun2021privacy} platforms have been recently proposed to secure sharing of medical data.
\par
Since the task of effectively anonymizing medical images is challenging, to the best of our knowledge, there is currently no way to achieve adequate anonymization of medical images.
In our method, since compressed images are not generated from the actual distribution of the original dataset, the generation of compressed images provides a possibility for visual anonymization of medical images.
A recent study has also theoretically demonstrated the role of dataset distillation in privacy protection~\cite{dong2022privacy}.
\begin{figure}[t]
        \centering
        \subfigure[]{
        \centering
        \includegraphics[width=4.0cm]{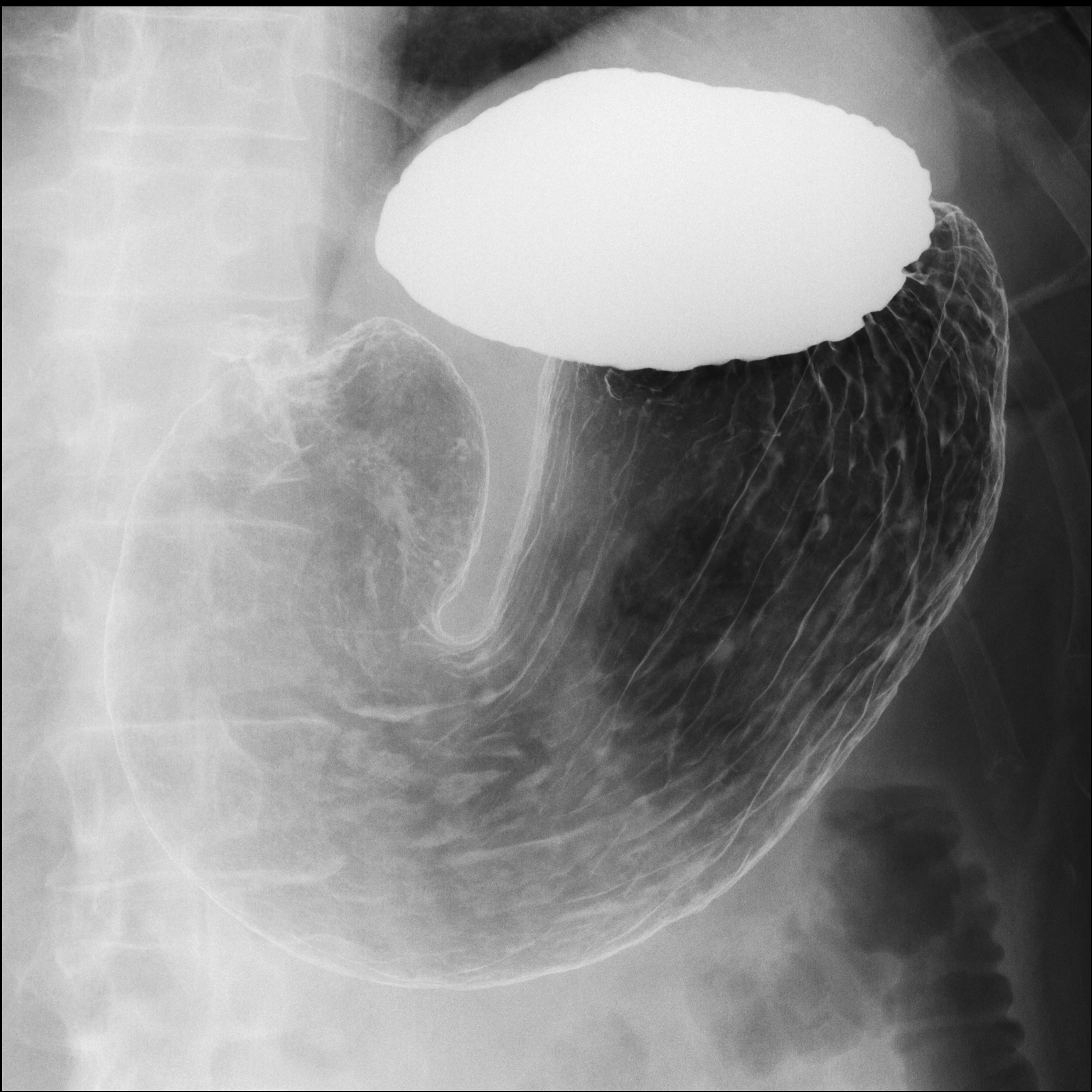}
        }
        \subfigure[]{
        \centering
        \includegraphics[width=4.0cm]{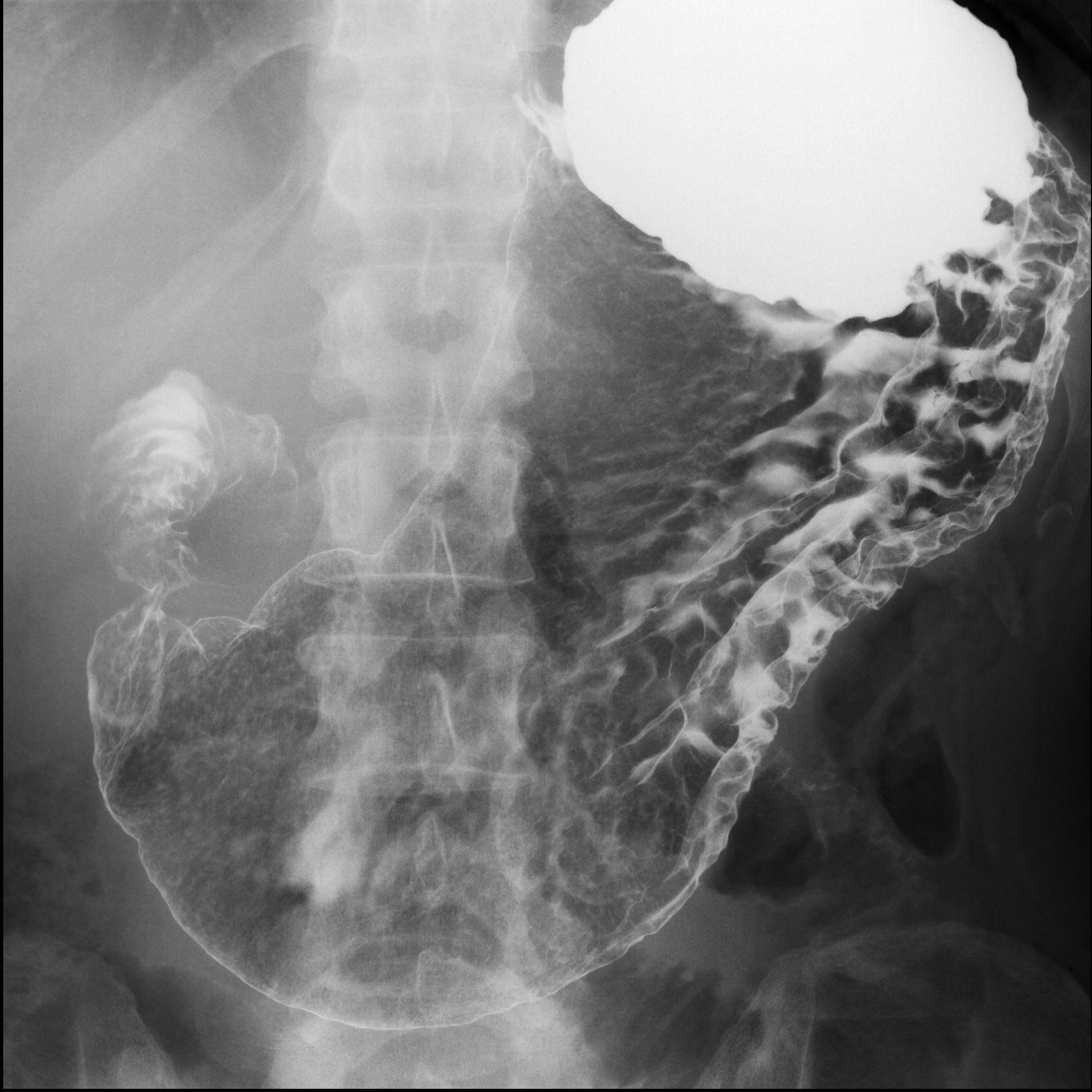}
        }
        \caption{Gastric images: (a) non-gastritis images and (b) gastritis images}
        \label{fig1}
\end{figure}
\begin{figure}[t]
        \centering
        \subfigure[]{
        \centering
        \includegraphics[width=2cm]{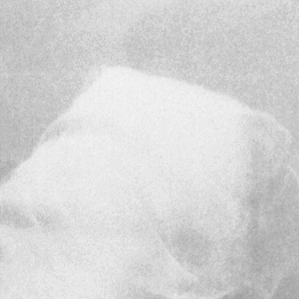}
        \includegraphics[width=2cm]{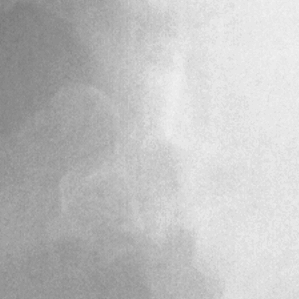}
        \includegraphics[width=2cm]{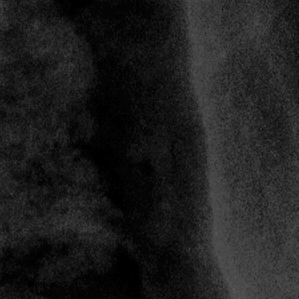}
        \includegraphics[width=2cm]{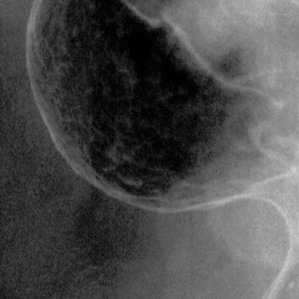}
        }
        \subfigure[]{
        \centering
        \includegraphics[width=2cm]{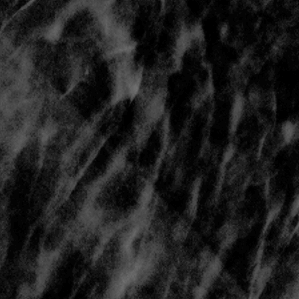}
        \includegraphics[width=2cm]{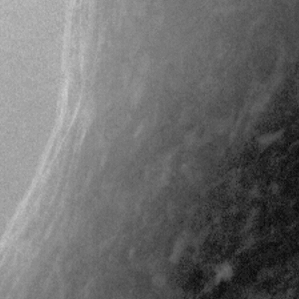}
        \includegraphics[width=2cm]{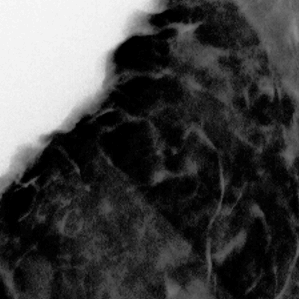}
        \includegraphics[width=2cm]{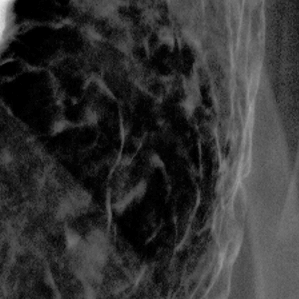}
        }
        \subfigure[]{
        \centering
        \includegraphics[width=2cm]{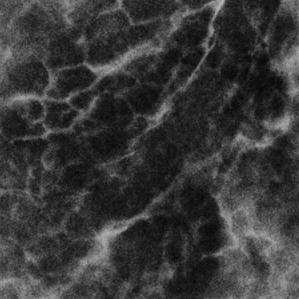}
        \includegraphics[width=2cm]{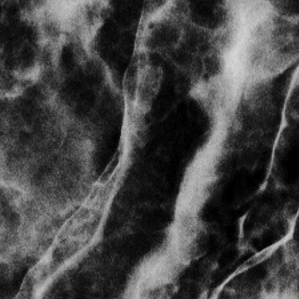}
        \includegraphics[width=2cm]{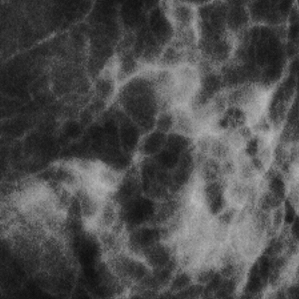}
        \includegraphics[width=2cm]{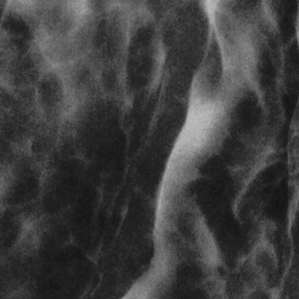}
        }
        \caption{Gastric patches: (a) irrelevant patches, (b) non-gastritis patches, and (c) gastritis patches}
        \label{fig2}
\end{figure}
\section{Proposed method}
\label{sec3}
This section presents the proposed method for generating compressed gastric images.
First, we explain the training data preprocessing procedure in Subsection~\ref{sec3-A}.
Then, we discuss the details of the compressed gastric image generation algorithm in Subsection~\ref{sec3-B}.
Finally, we show how to perform gastritis detection with patches in Subsection~\ref{sec3-C}.
\subsection{Training data preprocessing}
\label{sec3-A}
In this subsection, we present a method to preprocess the training data considering clinical settings. 
Figure~\ref{fig1} shows the full gastric X-ray images used in this study: (a) is an example without gastritis (hereafter called non-gastritis), and (b) is an example with gastritis.
As shown in Figure~\ref{fig1}, a stomach without gastritis has straight folds and uniform mucosal surface patterns. 
However, a stomach with gastritis has non-straight folds and coarse mucosal surface patterns.
The gastric images have high resolutions (i.e., 2,048 $\times$ 2,048 pixels).
In practical medical applications, high-resolution images could lead to expensive computational costs.
Since patch-based detection methods can efficiently use information about pathological regions and locations and reduce the computational cost, we follow our previous approaches to divide each image into patches for compressed gastric image generation~\cite{li2020soft, li2020complexity}.
\par
First, we divide each gastric X-ray image into multiple patches.
Let $X_{\mathrm{train}} \in \mathbb{R}^{d\times d}$ represent a gastric image in training data.
The label of $X_{\mathrm{train}}$ is defined as $Y_{\mathrm{train}} \in \{0, 1\}$, where $Y_{\mathrm{train}} = 0$ and $Y_{\mathrm{train}} = 1$ represent non-gastritis and gastritis, respectively.
Specifically, the full gastric images are divided into $H \times W$ patches ($H$ and $W$ are the numbers of patches in the vertical and horizontal directions, respectively).
We define the divided patches as $( \mathbf{x, y} )$ = $\{ x_{g}, y_{g} \}_{g=1}^{G}$, where $G$ represents the number of patch images, $x_{g} \in \mathbb{R}^{d'\times d'}$ represents the image and $y_{g}$ represents its label. 
Next, we annotate the patch images into the following three categories (i.e., $y_{g} \in \{\mathcal{I}, \mathcal{N},\mathcal{P}\}$): 
\begin{itemize}
    \item $\mathcal{I}$: patches extracted from the outside of the stomach (irrelevant) in gastritis and non-gastritis X-ray images, 
    \item $\mathcal{N}$: patches extracted from the inside of the stomach in non-gastritis (negative) X-ray images,
    \item $\mathcal{P}$: patches extracted from the inside of the stomach in gastritis (positive) X-ray images. 
\end{itemize}
A radiological technologist manually processed the stomach region annotation in this research. 
Figure~\ref{fig2} shows some examples of patch images.
We used the divided patches for the following compressed gastric image generation.
\subsection{Compressed gastric image generation}
\label{sec3-B}
This section explains how to generate compressed gastric images based on soft-label dataset distillation.
Since the training and test phases of dataset distillation differ from the standard neural network training and testing, we first briefly introduce the framework of the proposed method.
In the training phase, we distill the information of a large dataset into several compressed anonymous images through a twice-differentiable loss and keep updating as the gradient descent.
In the test phase, we use the optimal distilled images to reproduce the accuracy of the training phase.
\par
\begin{algorithm}[t]
    \caption{Training phase}    
    \label{alg1}
    \begin{algorithmic}[1]
    \REQUIRE 
    $\theta$: the random initial weights of a DCNN model;
    $\alpha$: learning rate;
    $K$: batch size;
    $T$: training steps;
    $M$: the number of compressed images;
    $\tilde{\mathbf{y}}_{0}$: initial value for $\tilde{\mathbf{y}}$;
    $\tilde{\alpha}_{0}$: initial value for $\tilde{\alpha}$
    \ENSURE
    $\tilde{\mathbf{x}}$: compressed images;
    $\tilde{\mathbf{y}}$: distilled labels;
    $\tilde{\alpha}$: optimized learning rate;
    $\theta_{\mathrm{bn}}$: batch normalization parameters 
    \\
    \STATE
    Initialize 
    $\tilde{\mathbf{x}}$ = $\{ \tilde{x}_{m} \}_{m=1}^{M}$ randomly,
    $\tilde{\mathbf{y}}$ = $\{ \tilde{y}_{m} \}_{m=1}^{M}\leftarrow\tilde{\mathbf{y}}_{0}$,
    $\tilde{\alpha}\leftarrow\tilde{\alpha}_{0}$
    \FOR{each training step $t = 1$ to $T$}
    \STATE
    Sample a mini-batch of training data:\\
    $( \mathbf{x}_{t}, \mathbf{y}_{t} )$ = $\{ x_{t,k}, y_{t,k} \}_{k=1}^{K}$
    \STATE
    Compute optimized weights using a gradient descent method:\\
    $\theta_{\mathrm{opt}} \leftarrow \theta - \tilde{\alpha} \, \nabla_{\theta} \ell ( \tilde{\mathbf{x}}, \tilde{\mathbf{y}}, \theta )$ 
    \STATE
    Evaluate the objective function on the mini-batch of training data:\\
    $\mathcal{L} = \ell ( \mathbf{x}_{t}, \mathbf{y}_{t}, \theta_{\mathrm{opt}} )$ 
    \STATE
    Update distilled data:\\
    $\tilde{\mathbf{x}} \leftarrow \tilde{\mathbf{x}} - \alpha \nabla_{\tilde{\mathbf{x}}} \mathcal{L}$,
    $\tilde{\mathbf{y}} \leftarrow \tilde{\mathbf{y}} - \alpha \nabla_{\tilde{\mathbf{y}}} \mathcal{L}$,
    and $\tilde{\mathbf{\alpha}} \leftarrow \tilde{\mathbf{\alpha}} - \alpha \nabla_{\tilde{\mathbf{\alpha}}} \mathcal{L}$
    \IF{the DCNN model has batch normalization layers}
    \STATE
    Save the batch normalization parameters as $\theta_{\mathrm{bn}}$
    \ENDIF
    \ENDFOR
    \end{algorithmic}
\end{algorithm}
Algorithm~\ref{alg1} shows the training phase of the proposed method.
First, we explain the settings of the training phase.
Here, $\theta$ represents the random initial weights of a random DCNN model; $\alpha$, $K$, and $T$ represent the learning rate, batch size, and training steps, respectively. 
$M$ represents the number of compressed images.
$\tilde{\mathbf{y}}_{0}$ is the initial value of distilled labels $\tilde{\mathbf{y}}$, and $\tilde{\alpha}_{0}$ is the initial value of optimized learning rate $\tilde{\alpha}$.
We can obtain the compressed images $\tilde{\mathbf{x}}$, distilled labels $\tilde{\mathbf{y}}$, optimized learning rate $\tilde{\alpha}$ and batch normalization parameters $\theta_{\mathrm{bn}}$ for the test phase.
\par
Next, we show the details of the compressed gastric image generation algorithm. 
For the training set of patches $ ( \mathbf{x, y} )$ = $\{ x_{g}, y_{g} \}_{g=1}^{G}$, $G$ represents the number of training images, and $x_{g}$ and $y_{g}$ represent the image and its label, respectively.
We define the weights of a random DCNN model as $\theta$. 
Let $\ell ( \mathbf{x, y}, \theta )$ be the loss of the DCNN model on the entire training set $ ( \mathbf{x, y} )$.
In our compressed gastric image generation method, we distill valid information of $ ( \mathbf{x, y} )$ to a very small distilled dataset $ ( \tilde{\mathbf{x}}, \tilde{\mathbf{y}} )$ = $\{ \tilde{x}_{m}, \tilde{y}_{m} \}_{m=1}^{M}$.
Here, $M$ represents the number of compressed images and $M \ll G$, $\tilde{x}_{m}$ and $\tilde{y}_{m}$ represent the distilled image and its distilled label, respectively.
\par
Since different categories of gastric patches may have common features, we let the compressed images $\tilde{\mathbf{x}}$ have soft labels $\tilde{\mathbf{y}}$, which can be represented with probability labels of $\mathcal{I}$, $\mathcal{N}$ and $\mathcal{P}$~\cite{hinton2014distilling, sucholutsky2019soft}.
The number of compressed images can be much smaller than the number of training images since the compressed images $\tilde{\mathbf{x}}$ do not come from the actual distribution.
Furthermore, soft labels play an essential role in the regularization during the training process, resulting in better detection performance than the original dataset distillation method~\cite{wang2018dataset}. 
The number of compressed images can be one image with soft labels, allowing for maximum compression of the gastric training set.  
Moreover, the optimized weights for the distilling process are calculated as follows:
\begin{equation}
\label{equ1}
\theta_{\mathrm{opt}} \leftarrow \theta - \tilde{\alpha} \nabla_{\theta} \ell ( \tilde{\mathbf{x}}, \tilde{\mathbf{y}}, \theta ),
\end{equation}
where $\ell ( \tilde{\mathbf{x}}, \tilde{\mathbf{y}}, \theta )$ represents the twice-differentiable loss of the DCNN model on the distilled dataset $ ( \tilde{\mathbf{x}}, \tilde{\mathbf{y}} )$, and $\tilde{\alpha}$ represents the optimized learning rate.
\par
Generally, the training of DCNNs is to find the minimizer of the empirical error over the entire training set $ ( \mathbf{x, y} )$ as follows:
\begin{equation}
\theta^{*}  = 
\mathrm{arg \, min} \, \ell ( \mathbf{x, y,} \, \theta ).
\end{equation}
Unlike the training goal of general DCNNs to find the optimal parameters $\theta^{*}$, our goal is to find the best-compressed images $\tilde{\mathbf{x}}^{*}$, distilled labels $\tilde{\mathbf{y}}^{*}$ and optimized learning rate $\tilde{\mathbf{\alpha}}^{*}$.
With the newly derived weights $\theta_{\mathrm{opt}}$ of the DCNN model, we can find the minimizer of the empirical error over the entire training set $ ( \mathbf{x, y} )$.
The objective function can be defined as follows:
\begin{equation}
\label{equ2}
\begin{split}
\tilde{\mathbf{x}}^{*}, \tilde{\mathbf{y}}^{*}, \tilde{\mathbf{\alpha}}^{*} & = 
\mathrm{arg \, min} \,\mathcal{L} ( \tilde{\mathbf{x}}, \tilde{\mathbf{y}}, \tilde{\alpha}; \theta ), \\ & = 
\mathrm{arg \, min} \, \ell ( \mathbf{x, y,} \, \theta_{\mathrm{opt}} ), \\ & =
\mathrm{arg \, min} \, \ell ( \mathbf{x, y,} \, \theta - \tilde{\alpha} \nabla_{\theta} \ell ( \tilde{\mathbf{x}}, \tilde{\mathbf{y}}, \theta ) ), 
\end{split}
\end{equation}
where $\mathcal{L}$ is differentiable and $\ell$ is twice-differentiable.
\par
To get the best compressed images, distilled labels, and optimized learning rate, we update the distilled data $\tilde{\mathbf{x}}$, $\tilde{\mathbf{y}}$, and $\tilde{\mathbf{\alpha}}$ at each distilling step using a gradient descent method as follows:
\begin{equation}
\label{equ3}
\begin{split}
&
\tilde{\mathbf{x}} \leftarrow \tilde{\mathbf{x}} - \alpha \nabla_{\tilde{\mathbf{x}}} \mathcal{L}, \\ &
\tilde{\mathbf{y}} \leftarrow \tilde{\mathbf{y}} - \alpha \nabla_{\tilde{\mathbf{y}}} \mathcal{L}, \\ &
\tilde{\mathbf{\alpha}} \leftarrow \tilde{\mathbf{\alpha}} - \alpha \nabla_{\tilde{\mathbf{\alpha}}} \mathcal{L},
\end{split}
\end{equation}
where $\nabla_{\tilde{\mathbf{x}}} \mathcal{L}$, $\nabla_{\tilde{\mathbf{y}}} \mathcal{L}$ and $\nabla_{\tilde{\mathbf{\alpha}}} \mathcal{L}$ represent the gradients of $\mathcal{L}$ based on $\tilde{\mathbf{x}}$, $\tilde{\mathbf{y}}$ and $\tilde{\mathbf{\alpha}}$, respectively, and $\alpha$ represents the learning rate.
\par
Then we explain the training process of the proposed algorithm.
First, we randomly initialize the compressed images $\tilde{\mathbf{x}}$.
Next, we initialize $\tilde{\mathbf{y}}$ and $\tilde{\alpha}$ with $\tilde{\mathbf{y}}_{0}$ and $\tilde{\alpha}_{0}$, respectively. 
At each training step $t$, we obtain a mini-batch of training data $( \mathbf{x}_{t}, \mathbf{y}_{t} )$ for which the size is $K$.
In the distilling process, we compute the optimized weights using Eq.~(\ref{equ1}). 
Additionally, we can extend the distilling process by performing multiple distill epochs and multiple distill steps for better distillation results.  
Specifically, we compute the optimized weights by performing sequential gradient descent steps on the distilled dataset and repeating a few epochs.
The calculation of sequential gradient descent steps is given as follows: 
\begin{equation}
\theta_{i+1} \leftarrow \theta_{i} - \tilde{\alpha} \nabla_{\theta_{i}} \ell ( \tilde{\mathbf{x}}, \tilde{\mathbf{y}}, \theta_{i} ),
\end{equation}
where $i$ represents the distill steps.
Then we evaluate the objective function on the mini-batch of training data using Eq.~(\ref{equ2}).
We update the distilled data $\tilde{\mathbf{x}}$, $\tilde{\mathbf{y}}$, and $\tilde{\alpha}$ using Eq.~(\ref{equ3}) based on a gradient descent method. 
Finally, we save the batch normalization parameters as $\theta_{\mathrm{bn}}$ if the DCNN model has batch normalization layers in its architecture. 
\par
\begin{algorithm}[t]
    \caption{Test phase}    
    \label{alg2}
    \begin{algorithmic}[1]
    \REQUIRE 
    $\theta$: the random initial weights of a DCNN model;
    $\tilde{\mathbf{x}}$: compressed images;
    $\tilde{\mathbf{y}}$: distilled labels;
    $\tilde{\alpha}$: optimized learning rate;
    $\theta_{\mathrm{bn}}$: batch normalization parameters 
    \ENSURE
    $\mathrm{Pred}$: predicted labels 
    \\
    \IF{the DCNN model does not have batch normalization layers}
    \STATE
    Compute optimized weights with the distilled data:\\
    $\theta_{\mathrm{opt}} \leftarrow \theta - \tilde{\alpha} \, \nabla_{\theta} \ell ( \tilde{\mathbf{x}}, \tilde{\mathbf{y}}, \theta )$
    \ELSE
    \STATE
    Compute optimized weights with the distilled data and batch normalization parameters:\\
    $\theta_{\mathrm{opt}} \leftarrow \theta - \tilde{\alpha} \, \nabla_{\theta} \ell ( \tilde{\mathbf{x}}, \tilde{\mathbf{y}}, \theta_{\mathrm{bn}}, \theta )$
    \ENDIF
    \STATE
    Predict the labels of the test data:\\
    $\mathrm{Pred}$ = $\mathrm{model} \, (\mathbf{x}_{\mathrm{test}}, \mathbf{y}_{\mathrm{test}}, \theta_{\mathrm{opt}})$
    \end{algorithmic}
\end{algorithm}
Note that the original dataset distillation method designed for simple datasets (e.g., MNIST and CIFAR10) only uses simple networks that do not contain batch normalization layers.
When a DCNN model has batch normalization layers, the mean and variance information of batches will be treated as constant and assumed not to change during the gradient steps in the distillation process.
Thus, the batches$'$ information cannot be distilled into compressed images.  
However, only the batch normalization parameters $\theta_{\mathrm{bn}}$ need to be saved to reproduce the training phase$'$s detection performance. This feature is used in our method to compress the sizes of trained models.
\par
Algorithm~\ref{alg2} shows the test phase of the proposed method.
In the test phase, if the DCNN model does not have batch normalization layers, we employ the distilled data $\tilde{\mathbf{x}}$, $\tilde{\mathbf{y}}$, and $\tilde{\alpha}$ to compute the optimized weights $\theta_{\mathrm{opt}}$ as follows:
\begin{equation}
\theta_{\mathrm{opt}} \leftarrow \theta - \tilde{\alpha} \, \nabla_{\theta} \ell ( \tilde{\mathbf{x}}, \tilde{\mathbf{y}}, \theta ).
\end{equation}
If the DCNN model has batch normalization layers, we can compute the optimized weights with the distilled data and batch normalization parameters as follows:
\begin{equation}
\theta_{\mathrm{opt}} \leftarrow \theta - \tilde{\alpha} \, \nabla_{\theta} \ell ( \tilde{\mathbf{x}}, \tilde{\mathbf{y}}, \theta_{\mathrm{bn}}, \theta ).
\end{equation}
Finally, we can predict the labels on the test data $(\mathbf{x}_{\mathrm{test}}, \mathbf{y}_{\mathrm{test}})$ with the obtained optimized weights $\theta_{\mathrm{opt}}$ as follows:
\begin{equation}
\mathrm{Pred} = \mathrm{model} \, (\mathbf{x}_{\mathrm{test}}, \mathbf{y}_{\mathrm{test}}, \theta_{\mathrm{opt}}),
\end{equation}
where $\mathrm{Pred}$ is the predicted labels of the test data and can be used for the final gastritis detection.
\subsection{Gastritis detection}
\label{sec3-C}
This subsection demonstrates how to perform gastritis detection with patches.
First, given a test gastric image $X_{\mathrm{test}} \in \mathbb{R}^{d\times d}$, it can be divided into $H \times W$ patches in the same manner as that for training data.
We can obtain the predicted labels (i.e., $\mathrm{Pred}$) of these patches by inputting the divided patches into a DCNN model with the optimized weights $\theta_{\mathrm{opt}}$.
Then, we calculate the numbers of patches whose predicted labels are $\mathcal{N}$ and $\mathcal{P}$ as $\mathrm{Num}(\mathcal{N})$ and $\mathrm{Num}(\mathcal{P})$, respectively.
Since patches $\mathcal{I}$ were extracted from outside the stomach and are not related to image-level ground truth, we do not use them for gastritis detection.  
Finally, we perform gastritis detection as follows:
\begin{equation}
\label{equ4}
Y_{\mathrm{test}} = 
\left\{\begin{matrix}
 1 & \mathrm{if} \, \frac{\mathrm{Num}(\mathcal{P})}{\mathrm{Num}(\mathcal{N})+\mathrm{Num}(\mathcal{P})} \geq \delta  \hfill \\
 0 & \mathrm{otherwise} \hfill
\end{matrix}\right.
,
\end{equation}
where $\delta$ is a threshold.
Note that when $Y_\mathrm{test} = 1$, the predicted result is gastritis, and the predicted result is non-gastritis when $Y_\mathrm{test} = 0$.
\section{Experiments}
\label{sec4}
In this section, we conduct three experiments to verify the effectiveness of the proposed method.
Subsection~\ref{sec4-A} presents the experimental settings.
The effectiveness of the dataset reduction and model compression of our method are presented in Subsections~\ref{sec4-B} and~\ref{sec4-C}, respectively.
Finally, we demonstrate the minimum number of compressed images of different DCNN models in Subsection~\ref{sec4-D}.
In all experiments, we implemented the algorithm using PyTorch~\cite{paszke2019pytorch} framework with an NVIDIA Tesla P100 GPU.
\subsection{Experimental settings}
\label{sec4-A}
The medical dataset used in our research contains gastric X-ray images for 815 patients (240 gastritis and 575 non-gastritis images).
The ground truth of each image was determined from patient diagnosis results of endoscopic and X-ray examinations. 
Gastric X-ray images were 2,048 $\times$ 2,048 pixels and gray-scale (8-bit). 
The training data contained images for 200 patients (100 gastritis and 100 non-gastritis images).
We used images for the remaining patients (140 gastritis and 475 non-gastritis images) as test data.
\par
In the data preprocessing stage, we divided all of these images into multiple patches (299 $\times$ 299 pixels) with a sliding interval of 50 pixels (i.e., $H$ = $W$ = 35) because we demonstrated in our previous study that good gastritis detection performance can be obtained in this case~\cite{kanai2019gastritis}.
For training data, a radiological technologist annotated the obtained patches as $\mathcal{I}$, $\mathcal{N}$, and $\mathcal{P}$.
If the area within the stomach is less than 1$\%$ in a patch, it is marked as $\mathcal{I}$.
Additionally, if the area within the stomach exceeds 85$\%$ in a patch, it is marked as $\mathcal{N}$ or $\mathcal{P}$.
We discarded the rest patches.
Consequently, we obtained training data $\mathcal{I}$, $\mathcal{N}$, and $\mathcal{P}$ for which the numbers of patches were 48,385, 42,785, and 45,127, respectively.
These patches were used to train DCNN models and for compressed gastric image generation.
For the test data, each of the remaining 615 gastric X-ray images was divided into 1,225 patches in the same manner as that of training data.
In the test phase, the detection result of a gastric image was determined using Eq.~(\ref{equ4}).
\par
Then, we conducted three experiments to evaluate the effectiveness of the proposed method.
We used the compressed images with the best detection performance in our experiments. We evaluated the detection performance of full gastric images in all three experiments.
In all experiments, we set the threshold $\delta$ to 0.4, which tended to have better detection performance.
The random initial weights $\theta$ of all DCNN models used in our experiments were initialized with the default Xavier initializer. 
The loss $\ell$ was a cross-entropy loss.
As in our previous study~\cite{togo2019detection}, we used sensitivity (Sen), specificity (Spe), and their harmonic mean (HM) as evaluation metrics, given as follows:
\\
\begin{equation}
\mathrm{Sen} = \frac{\mathrm{TP}}{\mathrm{TP + FN}},
\end{equation}
\begin{equation}
\mathrm{Spe} = \frac{\mathrm{TN}}{\mathrm{TN + FP}},
\end{equation}
\begin{equation}
\mathrm{HM} = \frac{\mathrm{2 \times Sen \times Spe}}{\mathrm{Sen + Spe}},
\end{equation}
\\
where TP, TN, FP, and FN represent the numbers of true positive, true negative, false positive, and false negative, respectively.
Since Sen and Spe have a tradeoff relationship, we took the HM as the final evaluation metric.
\subsection{Demonstration of the effectiveness of dataset reduction}
\label{sec4-B}
This subsection demonstrates the dataset reduction effectiveness of the proposed method through a comparison of dataset distillation and general network training.
We used ResNet18~\cite{he2016deep} and VGG16~\cite{simonyan2015very}.
First, we set the number of compressed images to 3 (one image per category) for the soft-label dataset and original dataset distillations~\cite{wang2018dataset} as SLDD (3) and DD (3), respectively.
Since the original dataset distillation method cannot distill the knowledge of the entire dataset into one compressed image, we only set the number of distilled images to 1 for soft-label dataset distillation as SLDD (1).
Then we initialized the soft labels with one-hot values of the original labels (i.e., $\mathcal{I}$, $\mathcal{N}$ and $\mathcal{P}$) in SLDD (3).
We also initialized the soft label with label $\mathcal{N}$ in SLDD (1), which tends to have better detection performance.
The distill epochs and steps were set to 3 (a total of nine distill steps).
DD (3) had the same settings as SLDD (3), except for the labels being fixed.
Consequently, we distilled all the training data into three images (one image per category) or one image. It saved the compressed images, distilled labels, and optimized the learning rate.
In the training phase, we performed 400 epochs in SLDD (3), DD (3), and SLDD (1) and saved the distilled results for testing and evaluation.
We randomly selected 1,000, 2,000, and 3,000 images per category and trained three ResNet18 models.
Furthermore, we use the following comparative methods (CMs) to prove the high-detection performance achieved using the proposed method.
\\
$\mathbf{CM1}$. We trained the VGG16 from scratch by considering the stomach regions (using all patches).
\\
$\mathbf{CM2}$. We fine-tuned the pre-trained VGG16 without considering the regions of the stomach (using only patches $\mathcal{N}$ and $\mathcal{P}$).
\\
$\mathbf{CM3}$. We trained the VGG16 from scratch without considering the regions of the stomach (using only patches $\mathcal{N}$ and $\mathcal{P}$).
\\
$\mathbf{CM4}$. We fine-tuned the VGG16 with the full gastric X-ray images, whose resolution is scaled down.
\\
$\mathbf{CM5}$~\cite{kanai2019gastritis}. We fine-tuned the pre-trained VGG16 by considering the stomach regions (using all patches).
\begin{table}[t]
    \caption{Test results of dataset distillation (ResNet18) and ResNet18.}
    \label{tab1}
    \begin{center}
    \begin{tabular}{lccc}
    \hline
    Method & Sen & Spe & HM \\\hline
    SLDD (3) & 0.886 & 0.869 & \bfseries{0.877} \\
    DD (3) & 0.829 & 0.884 & 0.856 \\
    ResNet18 (9000) & 0.814 & 0.832 & 0.823 \\
    ResNet18 (6000) & 0.907 & 0.760 & 0.827 \\
    ResNet18 (3000) & 0.914 & 0.669 & 0.773 \\\hline\hline
    SLDD (1) & 0.793 & 0.895 & 0.841 \\\hline
    \end{tabular}
    \end{center}
\end{table}
\begin{table}[t]
    \caption{Test results of dataset distillation (VGG16) and CMs.}
    \label{tab1-}
    \begin{center}
    \begin{tabular}{lccc}
    \hline
    Method & Sen & Spe & HM \\\hline
    SLDD (2) & 0.921 & 0.926 & \bfseries{0.923} \\
    CM1 & 0.950 & 0.886 & 0.917 \\
    CM2 & 0.907 & 0.891 & 0.899 \\
    CM3 & 0.829 & 0.813 & 0.821 \\
    CM4 & 0.700 & 0.705 & 0.703 \\\hline\hline
    CM5 & 0.964 & 0.945 & 0.955 \\\hline
    \end{tabular}
    \end{center}
\end{table}
\begin{figure}[t]
        \centering
        \includegraphics[width=8cm]{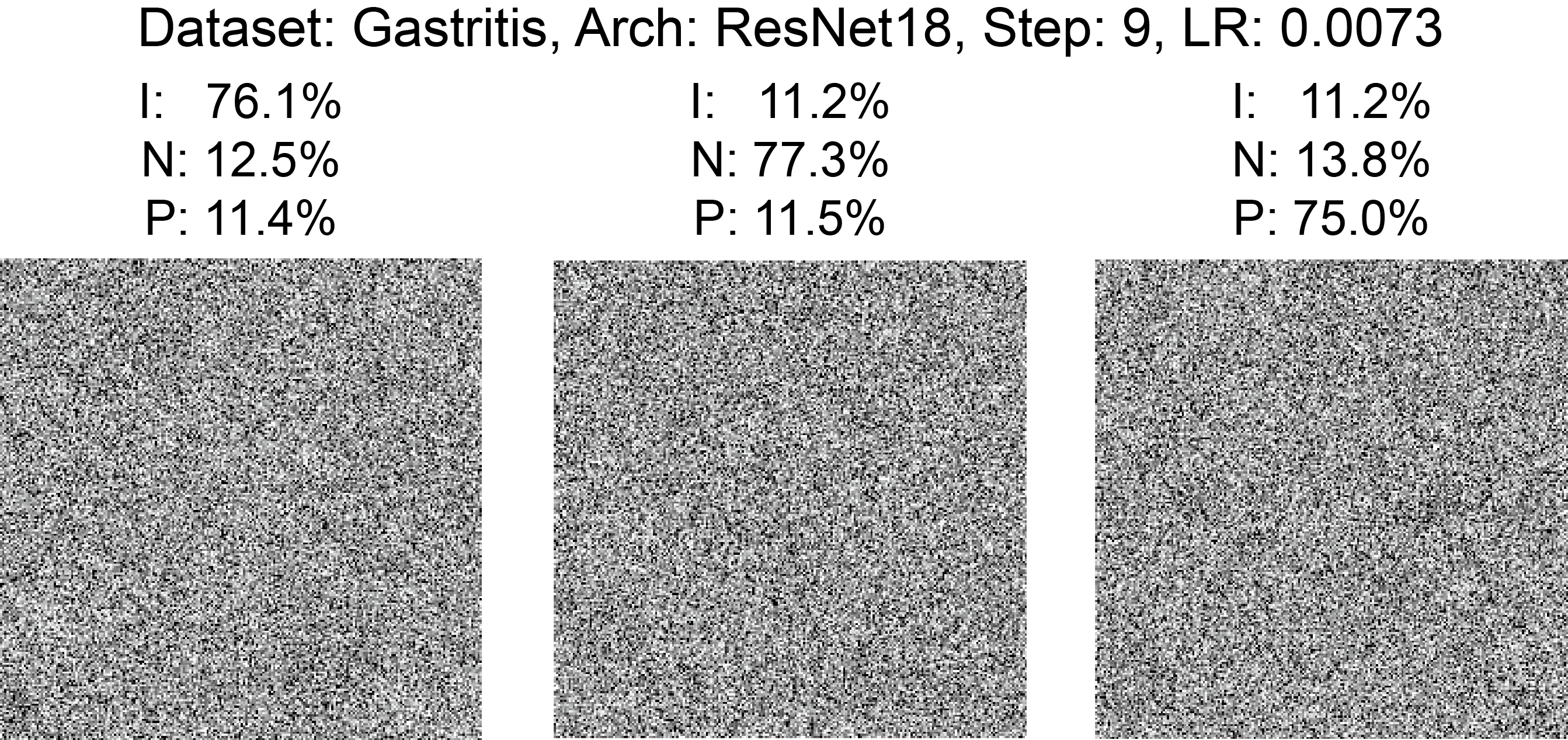}
        \caption{Compressed image generated in SLDD (3).}
        \label{fig3}
\end{figure}
\begin{figure}[t]
        \centering
        \includegraphics[width=8cm]{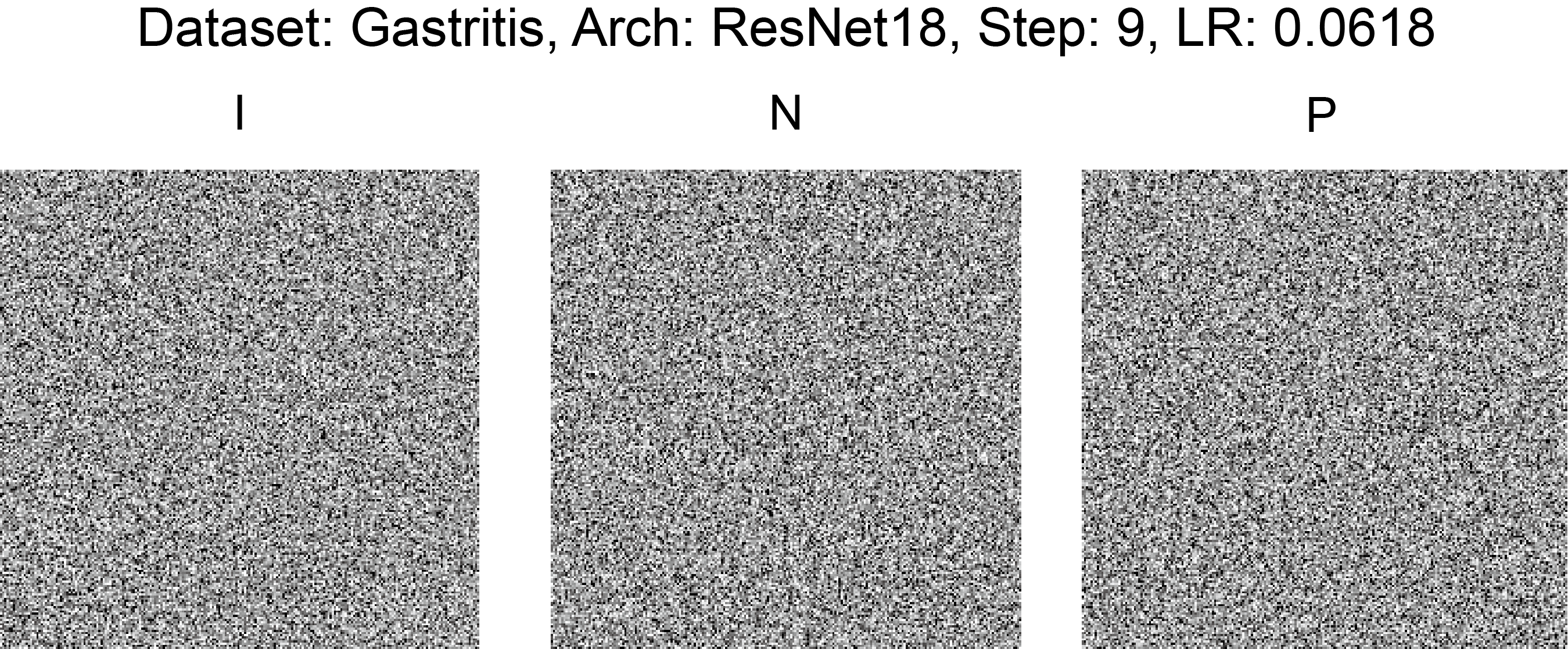}
        \caption{Compressed image generated in DD (3).}
        \label{fig4}
\end{figure}
\begin{figure}[t]
        \centering
        \includegraphics[width=6.7cm]{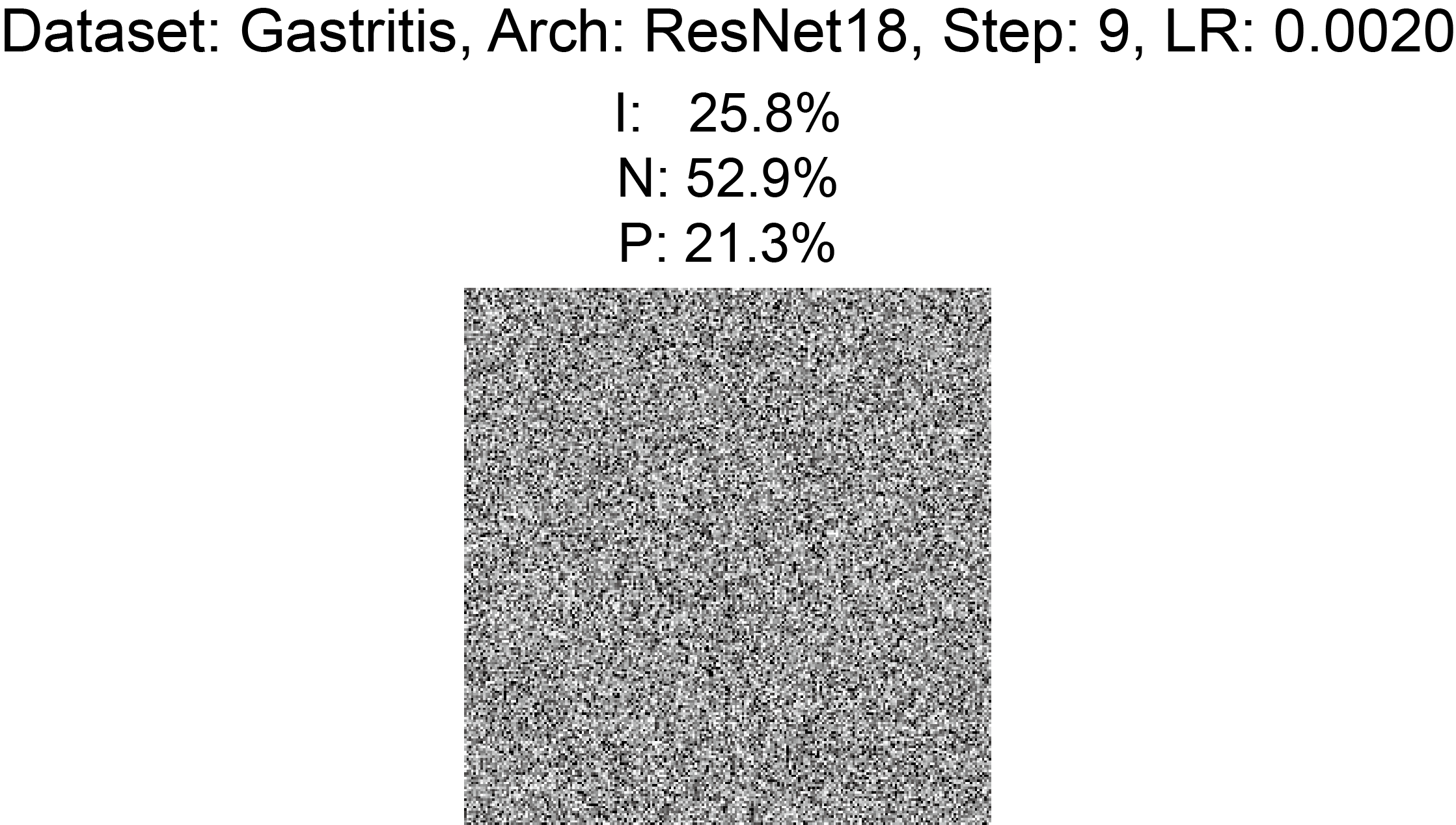}
        \caption{Compressed image generated in SLDD (1).}
        \label{fig5}
\end{figure}
\par
The test results are presented in Tables~\ref{tab1} and~\ref{tab1-}.
Table~\ref{tab1} presents the gastritis detection performance of the proposed method and ResNet18.
The ResNet18 model trained with 9,000 images (3,000 images per category) has an HM score of 0.823.
However, SLDD (1), which distilled all training data into only one compressed soft-label patch image for training, has an HM score of 0.841.
Furthermore, SLDD (3), which distilled all training data into three compressed soft-label images for training, has a higher HM score (0.877) than that DD (3).
We can see that a high-detection performance with several compressed gastric X-ray images, indicating the dataset reduction effectiveness of the proposed method.
The experimental results show that we can distill information of the entire gastric X-ray image dataset into only one compressed soft-label patch image for the maximum compression rate compared to our previous work~\cite{li2020soft}.
Table~\ref{tab1-} presents the gastritis detection performance of the proposed method and CMs trained on the full dataset.
SLDD (2), which distilled all training data into two compressed soft-label images for training, achieved comparable detection performance with CMs trained on the full dataset.
Figures~\ref{fig3},~\ref{fig4}, and~\ref{fig5} show examples of compressed images used in our experiments.
As shown in Figures~\ref{fig3},~\ref{fig4}, and~\ref{fig5}, the gastric images are anonymized visually.
Thus, the compressed patch images have no private information of patients and help privacy protection of medical data sharing.
\subsection{Demonstration of the effectiveness of model compression}
\label{sec4-C}
\begin{table}[t]
    \caption{Test results of different models with batch normalization (bn) and without batch normalization (no\_bn).}
    \label{tab2}
    \begin{center}
    \begin{tabular}{lccc}
    \hline
    Model & Sen & Spe & HM \\\hline
    GoogLeNet (bn) & 0.850 & 0.916 & \bfseries{0.882}\\
    GoogLeNet (no\_bn) & 0.121 & 0.823 & 0.211 \\\hline
    ResNet18 (bn) & 0.836 & 0.905 & \bfseries{0.869} \\
    ResNet18 (no\_bn) & 0.600 & 0.844 & 0.701 \\\hline
    AlexNet (bn) & 0.793 & 0.884 & \bfseries{0.836} \\
    AlexNet (no\_bn) & 0.786 & 0.861 & 0.822 \\\hline
    VGG16 (bn) & 0.907 & 0.926 & \bfseries{0.916} \\
    VGG16 (no\_bn) & 0.936 & 0.897 & \bfseries{0.916} \\\hline
    \end{tabular}
    \end{center}
\end{table}
This section demonstrates the model compression effectiveness of the proposed method through a comparison of different DCNN models with batch normalization layers or not.
Generally, when a DCNN model has batch normalization layers in its architecture, the information of a batch will be saved in the parameters.
In the training phase, the mean and variance of batches will be treated as constant and assumed not to change during the gradient steps, and hence the batches$'$ information cannot be distilled into compressed images.
However, only the batch normalization parameters need to be saved to reproduce the detection performance of the training phase.
Therefore, we can use this feature to compress the model$'$s size to be saved.
\par
We used different models (e.g., GoogLeNet~\cite{szegedy2015going}, ResNet18,  AlexNet~\cite{krizhevsky2012imagenet}, and VGG16) with and without batch normalization layers.
First, we set the number of compressed images to 3 (one image per category).
Then, we initialized the soft labels with one-hot values of the original labels.
The distill epochs and steps were set to 1.
In the training phase, we performed 400 epochs for GoogLeNet, ResNet18, and AlexNet, and 200 epochs for VGG16.
We saved the distilled results and the batch normalization parameters after every epoch for testing and evaluation.
\par
The test results are presented in Table~\ref{tab2}.
Table~\ref{tab2} shows that the models with batch normalization layers have better detection performance.
We also conducted experiments on multiple models, and the maximum compression rates of different models are presented in Table~\ref{tab3}.
Memory represents the memory required to save all of the parameters of different DCNN models.
Memory$\ast$ represents the memory required to save batch normalization parameters and distilled results of different models.
As presented in Table~\ref{tab3}, the proposed method can effectively reduce the memory size required to save the trained models, demonstrating the model compression effectiveness of the proposed method.
For example, VGG19 can have a maximum compression rate of 0.00076 with two compressed soft-label patch images.
We also found that the minimum number of compressed images that different models can achieve is not the same, and this feature is related to the maximum compression rate.
Thus, the next subsection presents the minimum number of compressed images of different models.
\begin{table}[t]
    \caption{Memory footprints of different models. Memory denotes saving all of the parameters of a model. Memory$\ast$ denotes saving batch normalization parameters and distilled results.}
    \begin{center}
    \label{tab3}
    \begin{tabular}{lccc}
    \hline
    Model & Memory & Memory$\ast$ & Compression rate \\\hline
    GoogLeNet & 22.83MB & 289.14KB & 0.01266 \\
    ResNet18 & 42.64MB & 250.23KB & 0.00587 \\
    ResNet34 & 81.20MB & 287.99KB & 0.00354 \\
    AlexNet & 217.44MB & 201.29KB & 0.00093 \\
    VGG16 & 512.21MB & 402.01KB & 0.00078 \\
    VGG19 & 532.46MB & 402.01KB & 0.00076\\\hline
    \end{tabular}
    \end{center}
\end{table}
\subsection{Minimum number of compressed images}
\label{sec4-D}
\begin{table}[t]
    \caption{The minimum number of compressed images of different models.}
    \label{tab4}
    \begin{center}
    \begin{tabular}{lccc}
    \hline
    Model & Sen & Spe & HM \\\hline
    GoogLeNet (1) & 0.764 & 0.853 & 0.806 \\
    GoogLeNet (3) & 0.850 & 0.916 & \bfseries{0.882} \\\hline
    ResNet18 (1) & 0.800 & 0.855 & 0.827 \\
    ResNet18 (3) & 0.836 & 0.905 & \bfseries{0.869} \\\hline
    ResNet34 (1) & 0.800 & 0.926 & 0.858 \\
    ResNet34 (3) & 0.893 & 0.899 & \bfseries{0.896} \\\hline
    AlexNet (1) & 0.671 & 0.895 & 0.767 \\
    AlexNet (3) & 0.793 & 0.884 & \bfseries{0.836} \\\hline
    VGG16 (1) & 0.643 & 0.524 & 0.577 \\
    VGG16 (2) & 0.921 & 0.926 & \bfseries{0.923} \\
    VGG16 (3) & 0.936 & 0.897 & 0.916 \\\hline
    VGG19 (1) & 0.614 & 0.891 & 0.727 \\
    VGG19 (2) & 0.921 & 0.909 & 0.915 \\
    VGG19 (3) & 0.921 & 0.933 & \bfseries{0.927} \\\hline
    \end{tabular}
    \end{center}
\end{table}
\begin{table}[t]
    \caption{Parameters of different models. Parameter denotes the number of model parameters. Image denotes the minimum number of compressed images.}
    \begin{center}
    \label{tab5}
    \begin{tabular}{lcc}
    \hline
    Model & Parameter & Image \\\hline
    GoogLeNet & 5,984,915 & 1 \\
    ResNet18 & 11,176,963 & 1 \\
    ResNet34 & 21,285,123 & 1 \\
    AlexNet & 57,000,643 & 1 \\
    VGG16 & 134,271,683 & 2 \\
    VGG19 & 139,581,379 & 2 \\\hline
    \end{tabular}
    \end{center}
\end{table}
This subsection presents the minimum number of compressed images of different DCNN models.
We found that the minimum number of compressed images that different models can achieve varies.
For example, when we set the number of compressed images to 1 with VGG16, the test accuracy was far lower than when the number of compressed images was set to 3 (one image per category).
In other words, VGG16 cannot effectively distill the valid information from the training data into one compressed image.
\par
We used different models, including GoogLeNet, ResNet18, ResNet34, AlexNet, VGG16, and VGG19. 
First, we set the numbers of compressed images to the minimum number of compressed images that the models can achieve, e.g., GoogLeNet (1) and VGG16 (2).
For comparison, we set the numbers of compressed images to 3 (one image per category), e.g., GoogLeNet (3).
The distill epochs and steps were set to 1.
In the training phase, we performed 400 epochs for GoogLeNet, ResNet18, and AlexNet, and 200 epochs for ResNet34, VGG16, and VGG19.
We saved the distilled results and the batch normalization parameters after every epoch for testing and evaluation.
\par
The test results are presented in Tables~\ref{tab4} and~\ref{tab5}. 
Table~\ref{tab4} shows that GoogLeNet, ResNet18, ResNet34, and AlexNet can effectively distill valid information from the training data into only one compressed soft-label patch image.
However, VGG16 and VGG19 cannot effectively distill valid information into one compressed patch image.
Since the compressed images were distilled with DCNN models, we think the minimum number of compressed images is related to the number of parameters.
Table~\ref{tab5} shows the parameters and the minimum number of compressed images of different models. 
The parameter represents the number of parameters of different models, and the image indicates the minimum number of compressed images that different models can achieve.
As presented in Table~\ref{tab5}, the minimum number of compressed images relates to the number of parameters.
Additionally, the minimum number of compressed images increases as the number of parameters of a DCNN model increases.
\section{Discussion}
\label{sec5}
In Subsection~\ref{sec4-B}, we showed that we could distill information of the entire gastric X-ray image dataset into only one compressed soft-label patch image for the maximum compression rate.
We achieved a high-detection performance with three compressed soft-label images for training the model.
Moreover, the visualization of the compressed soft-label images showed that they had been visually anonymized.
Subsection~\ref{sec4-C} presents the results of evaluating the performances of different DCNN models to verify the robustness of the proposed method.
We also showed the effectiveness of the batch normalization layers of the proposed method and how to reduce the memory required to save trained models.
Furthermore, Subsection~\ref{sec4-D} presents the relationships between the minimum number of compressed images of different DCNN models and their parameters.
\par
The experimental results showed that our method has limitations.
The theoretical proof of the minimum number of compressed images and the number of parameters is complex, and there is no work in this area. 
Therefore, we only make reasonable inferences and heuristics based on observed phenomena in this paper and leave it for future work.
We obtained that the minimum number of compressed images increases as the number of parameters of a DCNN model increases. 
Additionally, the complexity of distillation significantly increases since our distillation process includes a large number of secondary gradient calculations.
Recently, more efficient alternatives to dataset distillation algorithms have emerged to condensate small datasets~\cite{zhao2020dataset, zhao2021dataset, cazenavette2022dataset, zhou2022dataset}~\footnote{https://github.com/Guang000/Awesome-Dataset-Distillation}. 
These algorithms can also be extended for effective anonymous medical data sharing, which will be one of our future works.
Although we focused on gastric X-ray image data, our method could also be applied to other medical images, e.g., endoscopic and CT images, which will be one of our future works.
Moreover, our recent studies on self-supervised learning~\cite{li2021self, li2021triplet, li2022self, li2022tribyol} can learn discriminative representations from different images without manually annotated labels, which fits well with dataset distillation algorithms. 
\section{Conclusion}
\label{sec6}
This paper proposes a compressed gastric image generation method based on the soft-label dataset distillation for efficient anonymous medical data sharing.
The proposed method compresses the entire medical dataset into only one compressed soft-label patch image and reduces the size of a trained model to a few hundredths of its original size.
The compressed images obtained after distillation have been visually anonymized; therefore, they do not contain private information of the patients.
Furthermore, the proposed method achieves high-detection performance with only a small number of compressed images.
The experimental results show that the proposed method can improve the efficiency and security of medical data sharing, which is helpful for future research at the intersection of computer science and biomedicine.
Although our study only tested on gastric X-ray image data, we expect the findings and methodology to be applied to other medical images.
\section*{Declaration of competing interest}
None declared.
\section*{Acknowledgments}
In this study, the medical data were provided by The University of Tokyo Hospital in Japan. We express our thanks to Nobutake Yamamichi of the Graduate School of Medicine, The University of Tokyo, and Katsuhiro Mabe of the Junpukai Health Maintenance Center.
This study was supported in part by AMED Grant Number JP21zf0127004, the Hokkaido University-Hitachi Collaborative Education and Research Support Program, and the MEXT Doctoral program for Data-Related InnoVation Expert Hokkaido University (D-DRIVE-HU) program. This  study was conducted on the Data Science Computing System of Education and Research Center for Mathematical and Data Science, Hokkaido University. 
\bibliographystyle{elsarticle-num}
\bibliography{CMPB}
\end{document}